# Detection, Recognition and Tracking of Moving Objects from Real-time Video via Visual Vocabulary Model and Species Inspired PSO

Kumar S. Ray, Anit Chakraborty, Sayandip Dutta

*Abstract*— In this paper, we address the basic problem of recognizing moving objects in video images using Visual Vocabulary model and Bag of Words and track our object of interest in the subsequent video frames using species inspired PSO. Initially, the shadow free images are obtained by background modelling followed by foreground modeling to extract the blobs of our object of interest. Subsequently, we train a cubic SVM with human body datasets in accordance with our domain of interest for recognition and tracking. During training, using the principle of Bag of Words we extract necessary features of certain domains and objects for classification. Subsequently, matching these feature sets with those of the extracted object blobs that are obtained by subtracting the shadow free background from the foreground, we detect successfully our object of interest from the test domain. The performance of the classification by cubic SVM is satisfactorily represented by confusion matrix and ROC curve reflecting the accuracy of each module. After classification, our object of interest is tracked in the test domain using species inspired PSO. By combining the adaptive learning tools with the efficient classification of description, we achieve optimum accuracy in recognition of the moving objects. We evaluate our algorithm benchmark datasets: iLIDS, VIVID, Walking2, Woman. Comparative analysis of our algorithm against the existing state-of-the-art trackers shows very satisfactory and competitive results.

## I. INTRODUCTION

EFFECTIVE recognition of objects for tracking in real-time video stream and processing of data involve integration of background modelling, shadow removal, foreground modelling and proper detection of objects. Recognition of the detected objects is done by extracting the features obtained from the principle of bag of words (BOW). Extracted feature sets are tracked in the successive test frames via species inspired PSO.

Although, various detection and tracking algorithms exist, still; object detection, recognition, effective tracking of feature sets and adoptability of handling occlusions and other noise are still a standing challenge in the field of computer vision. In order to perform well, in any domain of interest, for any tracking algorithm, satisfactory training model is of utmost importance. In recent years, several attentions [51, 49, 43, 38, 32, 30, 17, 18, 20] have been given in this direction to achieve and share the goal of this paper. Generally, appearance based tracking algorithms are of two types mainly; generative [51, 32, 20, 18] and discriminative [49, 43, 38, 30, 17].

Several tracking algorithms based on static appearance models exists, which are either trained using only the first few consecutive set of iterations or defined manually [56], [36], [7], [35]. These algorithmic frameworks often fail to deal with momentous appearance changes, which cause nonlinear transformation of the appearance of a given object. Such challenges lead to difficulty when there is an absence of sufficient amount of priori knowledge. In our paper, we have considered several appearances of a given training object of our interest, which can capture the momentous and nonlinear changes of the appearances, which are explained in length in Section III and IV.

Here, we use the Visual Vocabulary Model using Bag of Words to extract the necessary features of certain instances of objects through rigorous high level training. Subsequently, we apply the features to the test datasets to recognize and locate our objects in the video scenes. Using visual instance occurrence and their probabilistic presence to imply a certain domain, we obtain optimum accuracy in domain recognition as well.

The contributions of this paper are:
- Background modelling and extraction of astute shadow free images using color invariant approach.
- Foreground modelling using morphological operators for effective reconstruction of the image.
- Adoptability of handling occlusions and other noises present in the test datasets.
- Extraction of the features of the objects captured in the blobs via the principle of Bag of Words.
- Classification of the objects in our domain of interest using probabilistic word occurrence for domain recognition.
- Tracking of the recognized objects via species inspired PSO.

Kumar S. Ray is with Indian Statistical Institute, 203 B.T Road, Kolkata 108, India. (e-mail: ksray@isical.ac.in)

Anit Chakraborty is with Indian Statistical Institute, 203 B. T Road, Kolkata – 108, India. (e-mail: ianitchakraborty@gmail.com).

Sayandip Dutta is with Indian Statistical Institute, 203 B.T Road, Kolkata 108, India. (e-mail: sayandip199309@gmail.com)

The organization of the paper constitutes: review of related works in the field of object detection and recognition, especially, based on supervised learning are briefly described in the following section (II). Section III explains the proposed method for detection, recognition and tracking, specifically, section III (D) describes the concept of Visual Vocabulary Model for object recognition. Experimental results on several datasets and the comparative analysis with some state-of-the-art algorithms are presented in section IV. Section V concludes the paper and discusses the future possibilities for further improvements.

## II. BRIEF REVIEW OF RELATED WORKS

Numerous color histograms based tracking algorithms [52, 55] have been proposed in recent years. The mean shift tracking algorithm has been extended by Collins [53] with the scale variation of object of interests in a video frame. Perez et al. [55] used color histogram in addition with a particle filter [57] for tracking of objects in video frames. A spatiogram based approach has been proposed to capture spatial relationships of statistical properties of pixel, in Birchfield and Rangarajan [45]. He et al. [7] developed a locality sensitive histogram at each pixel for finer distribution of the visual feature points for object tracking in video scenes. Histograms of oriented gradients (HOGs) [42] is proposed for object tracking [34] in addition to the integral histogram [44]. Covariance region descriptors [39] based approaches were introduced for tracking, to combine different features. Local binary patterns (LBP) [54] as well as Haar-like features [47] have been proposed for appearance based tracking of objects [17], [38], [19], [9]. Spatio-temporal representation combined with genetic algorithm has also been used for feature extraction [1]. Recently pixel based segmentations have been applied [2] to handle tracking.

Various generative models have been proposed for multiple object tracking in past years. In [58] and [59], Sparse Generative Appearance Modeling is implemented to build an appearance model of objects. Gaussian Mixture Models (GMM) [60] [62], are popular generative approach for tracking. Apart from GMM, several other mixture models have been used in tracking in earlier days, such as finite mixture models [63-68]. Priebe et al. [67] introduced an algorithm based on recursive mixture density estimation. To extract time-invariant characteristics, the authors of [69] present a Bayesian Tracking approach using autoregressive Hidden Markov Model (AR-HMM) for robust visual tracking.

In recent years, discriminative models are leading the way in the field of object tracking [49], [33]. In this method, a binary classifier is trained from the input video sequence, for separation of target and background. The classifiers that have been extensively used for object tracking are: ranking SVM [18], semi-boosting [30], support vector machine (SVM) [49], boosting [38], structured output SVM [19], and online multi-instance boosting [17]. In [49], a trained SVM classifier is integrated for tracking, to tackle appearance based changes with varying illumination.

A confidence map [43] in each frame is built using a discriminative feature combination, learned online, for separation of background from target objects. Larese et al. [3] have used SIFT descriptor to discriminate the patterns and then used it to build Bag of Words model and finally classified with SVM.

Various tracking codes are available for evaluation with significant effort of the authors, e.g., MIL [17], OAB [38], IVT [32], L1 [26], TLD [23] and likes.

## III. PROPOSED METHOD

Accurate detection of the objects of interest across multiple frames and tracking of the recognized objects are still a challenge. In order to do that, first we model the background and remove the hard shadows from the background to extract the exact area occupied by the object in a frame. Next, we model the foreground and subtract the background model without shadow to obtain the blob of an object. Before recognizing the object inside the blob, we train a machine learning inspired Visual Vocabulary Model with a set of objects which can represent our domain of interest for recognition and tracking. In both the cases, i.e., in case of training stage we extract the features of the objects of the training data by principle of Bag of Words and in the testing phase, we use the same technique to extract the features of the objects of the test data. After classification of extracted feature sets, we track the features of interest of the recognized objects in successive video frames via Species inspired PSO.

### A. Background Modeling

Background modelling is an integral part of our algorithm. This part consists of pretreatment of each frame of the video sequence, followed by temporal image analysis. We update the background continuously so as to adapt to the changing background and other variations. We set a background adaptive threshold in order to differentiate between foreground and background objects. Finally, we apply certain domain centric morphological operators to the updated background to obtain smooth background images.

In [31], Li *et al.* proposed an idea for background modelling. In our work, we introduced some modification of the same work and proceed as follows:

At each time step an image $I_m^t$ is obtained by subtracting two successive video frames and $F_m^t$ can be obtained by subtracting the current video frame with the background model. To deal with sudden illumination variation an AND-OR operation is performed over $I_m^t$ and $F_m^t$. (Fig. 1)

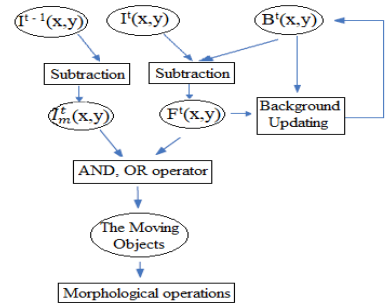

Fig. 1. Block diagram of the motion detection module. [31]

## a. Pretreatment

Transformation of color image to grayscale is defined as:

$$gray = 0.299R + 0.587G + 0.114B. \tag{1}$$

## b. Temporal image analysis:

The extracted frame $I^t$ is compared with its previous frame $I^{t-1}$ in order to obtain $I_m^t$ by predicting the similarity between the two consecutive pixel values of frames $I_t$ (x, y) and $I_{t-1}$(x, y), expressed using radiometric similarity R $(I^t$ (x,y), $I^{t-1}$ (x,y)) [31]:

$$R(I^t(x,y), I^{t-1}(x,y)) = \frac{E[W(I^t(x,y)W(I^{t-1}(x,y))] - E[W(I^t(x,y)E[W(I^{t-1}(x,y))]}{\sqrt{D[W(I^t(x,y)]D[W(I^{t-1}(x,y))]}}. \tag{2}$$

Mean and variance of the pixel intensities captured in a particular window of a specific video frame W; E[W], D[W]. Pixel centers are compared between the succeeding images $(I^t(x,y)$, $I^{t-1}(x,y))$.

Temporal binary image of the moving object $(I_m)$ has a radiometric similarity value, formally expressed as:

$$I_m(x,y) = \begin{cases} 1, & if\ R(x,y) > T_b \\ 0, & otherwise \end{cases} \tag{3}$$

Similarly, $F_m^t$ is formulated on a hypothesis based on the difference threshold $(T_b)$, between background frame and the current frame, formally:

$$F_m^t = \begin{cases} 1, & if\ |I^t(x,y) - B^t(x,y)| > T_b \\ 0, & otherwise \end{cases} \tag{4}$$

The pixels (x,y) of moving objects are formulated by operating on $I_m(x,y)$ and $F^t(x,y)$:

$$M^t(x,y) = \begin{cases} 1, & if\ (I_m(x,y) \cap F^t(x,y)) = 1) \\ 0, & otherwise \end{cases} \tag{5}$$

The moving pixels in video frames are identified by $M^t(x,y)$.

## c. Background updating

Background model is updated with newly arrived information from previous frame using the first-order recursive filter:

$$B^{t+1}(x,y) = B^t(x,y) + \alpha \times (I^t(x,y) - B^t(x,y)), \tag{6}$$

where $\alpha$ is an arbitrary adaptation coefficient.

In our implementation, a vector history V, with the six last values updated cumulatively, is considered as:

$$V = [E(t), E(t-1), E(t-2), E(t-3), E(t-5)]. \tag{7}$$

At time t, the mean value of the frame is E(t).

For each frame, we calculate proper learning rate $\alpha$, based on this vector:

$$\alpha = a + b \frac{|E(t) - E(t-5)|}{\max(E(t), E(t-5))}, \tag{8}$$

Typically, $a$ ranges from 0.04 to 0.06. Here $a$ is chosen to be 0.05. For a given value of $b$ and the gain as stated by $\frac{|E(t)-E(t-5)|}{\max(E(t),E(t-5))}$, usually we obtain a small value of $\alpha$. The slope of the gain variation curve, is denoted by $b$, where the gain is represented by $[\frac{|E(t)-E(t-5)|}{\max(E(t),E(t-5))}]$.

## d. Adaptive threshold:

The formulation of the noise is described as follows:

$$p(n) = \frac{1}{\sqrt{2\pi}\sigma} exp\left\{-\frac{(n-\mu)^2}{2\sigma^2}\right\}, \tag{9}$$

where p(n) is the probability of the background pixel.

Let $d$ be a pixel of the image, the gray histogram of the pixel is $h(d)$, and background pixels and foreground pixels are denoted by $I_B$ and $I_F$ respectively. Probability of a background pixel misidentified as foreground pixel and vice versa are as follows:

$$P_{F|B} = \sum_{d \in I_F} p(d \mid B) \text{ and } P_{B|F} = \sum_{d \in I_B} p(d \mid F), \tag{10}$$

where $P_{d|B}$ is the probability of background pixel and $P_{d|F}$ is the probability of foreground pixel.

Our goal is to minimize $P_{d|B}$ and $P_{d|F}$ as much as possible. The Min $P_{F|B}$ is significant, as after morphological operation in the post-process, $P_{B|F}$ will be smaller.

$p(B)$ is the priori probability of the background as calculated from gray histogram of the image $I_m^t$.

$$p(B) = \sum_{d=-T}^{T} h(d) \qquad \mu = 0. \tag{11}$$

Following is the noise of the variance model, as expressed:

$$\sigma = \sum_{d=-T}^{T} h^2(d)/p(B). \tag{12}$$

A fitting criterion describes the threshold value defined as,

$$e_{Min} = \sum_{d \in D}(p(B)p(d|B) - h(d))^2 . \tag{13}$$

## e. Morphologic image operation:

Apart from moving objects, binary images contain a lot of residual noise. The object detection fails fatally because of many pixel holes in the image. The morphological operators help to remove the holes present and many of the noise by smoothening the edges of the blob. The morphological operators used in this experiment are Adaptive Kernel operator, Gaussian operator and Laplacian Filter.

### B. Shadow removal

As mentioned in [41] by Xu *et al.*, the shadow removal approach is pictorially described in Fig. 2.

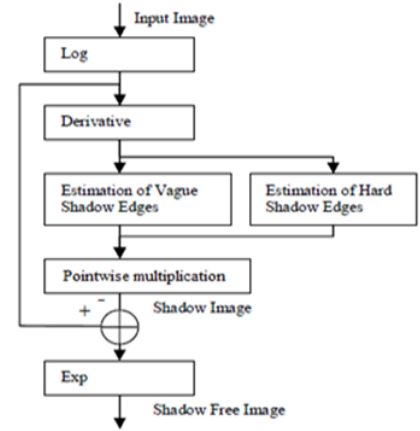

Fig. 2. Block diagram of the shadow removal system. [41]

The r,g,b normalization is formulated as follows:

$$r' = \frac{r}{\sqrt{r^2 + g^2 + b^2}}, \tag{14}$$

$$g' = \frac{g}{\sqrt{r^2 + g^2 + b^2}}, \tag{15}$$

$$b' = \frac{b}{\sqrt{r^2 + g^2 + b^2}}, \tag{16}$$

where r, g, b are input image color channels, r',b',g' construct the shadow-free color invariant image.

Application of Gaussian smooth filter suppresses the high frequency textures in both invariant and original images. These smoothed color images are converted to gray scale, following the HSV color model definition, to detect the edges, formally:

$$E_{ori} = \|edge(I_{ori})\|, E_{inv(i)} = \|edge(I_{inv(i)})\|, \quad (17)$$

where $E_{ori}$ is the edge of the original image after applying smooth filter and $I_{ori}$ is the original image. $E_{inv(i)}$ is the edge of the color invariant image after applying smooth filter and $I_{inv(i)}$ is the color invariant image.

Here, the values of $i$ (i.e. the invariant image index) are 1 and 2. The hard shadow edge mask is constructed by choosing the strong edges of original images that are absent in the invariant images. Thus, we get:

$$HS(x,y) = \begin{cases} 1, & E_{ori}(x,y) > t1, \& \\ & \min_i(E_{inv(i)}(x,y) < t2) \ , \\ 0, & otherwise \end{cases} \quad (18)$$

where $t1$, $t2$ are thresholds, set manually, and assessed hard shadow edge mask is $HS(x,y)$. In (18), $t1$ maps the selected shadow edges to the strong edges of the subsequent hard shadows in images. $t2$ selects edges belonging only to shadows.

### C. Foreground Modelling and Reconstruction

In the present work, we have introduced a new concept of foreground modeling. This is essential to recognize and detect the objects from static or moving background variants.

We consider Poisson equation solution to reinforce the shadow portions from the derivative frames. To begin with, shadow edge masks of the two kinds and are merged as:

$$mask = VS \mid HS, \quad (19)$$

where $VS$ represents vague shadow and $HS$ represents hard shadow.

Furthermore, masking is applied to the gradient field:

$$\nabla_{i'} = mask \cdot \nabla_i = (mask \cdot G_x \, , mask \cdot G_y), \quad (20)$$

where $G_x$ and $G_y$ are gradient field along the axis.

The clipped derivatives are denoted as Gx' and Gy' and the calculation of the scalar is represented as:

$$divG = \frac{\partial G'_x}{\partial x} + \frac{\partial G'_y}{\partial y}. \quad (21)$$

Finally, shadow image restoration is done by working out the well-known Poisson equation:

$$\nabla^2 s = divG, \quad (22)$$

where, $s$ is the calculated shadow image. As we perform the process in logarithmic domain, an exponent operation is added to the image:

$$S(x,y) = \exp(s(x,y) - \max_{x,y}(s(x,y))). \quad (23)$$

Shadow extraction from original image, as formulated:

$$r(x,y) = i(x,y) - s(x,y), \quad (24)$$
$$R(x,y) = \exp(r(x,y) - \max_{x,y}(r(x,y))), \quad (25)$$

where $S$ and $R$ are the shadow image and shadow free image respectively, after reconstruction.

Figure 3. (a) shows the actual video frames; column (b) is the modelled background, column (c) corresponds to the foreground modelling. The figures visible are accepted as foreground objects.

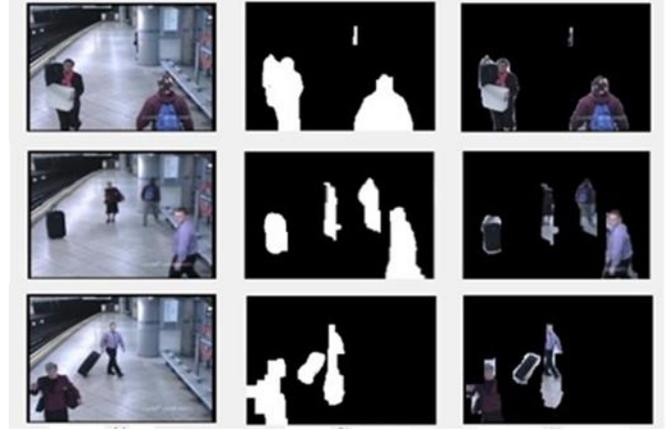

Fig. 3. (a) Video Frame    (b) Background Model    (c)Foreground Model

### D. Visual Vocabulary Model for Object Recognition

Visual Vocabulary Model is a machine learning based image classification model, specifically, handling images as documents, by labelling specific features as words. By observing presence of such feature key words in an image, Visual Vocabulary Model can predict the domain of the image and by their arrangement in an image, it can recognize different objects.

We have adopted this method as it yields satisfactory results with limited training samples, and the accuracy only increases with the increase in training datasets. In our paper, scene recognition is essential to narrow down the objects that can be present in a scene.

Recognition using Visual Vocabulary consists of selection of distinct key words and normalization of the surrounding regional content. Assigning a descriptor to the normalized content, helps matching the captured objects which are of our domain of interest.

#### a. Visual keyword localization:

To localize the keywords, first step is to extract the features of the object of interest such that they are distinct and invariant under different scale and illumination based conditions even with the presence of noise.

To construct the Visual Vocabulary a corpus of training images need to be stored in the feature space in order to model the descriptor instances. Subsequently, the modelled descriptors have to be clustered for quantization of the feature space into distinctive visual words, where the visual words signify the center of the cluster. For given video frames, the closest matching visual key points are recognized with the corresponding features. The bag of words feature sets can be used to define the principle description of any given videoframe. This process can be divided into three steps, namely; tokenization, counting and normalization. In tokenization phase, similar feature patches are labelled via k-mean clustering. In the counting phrase, the number of tokens are counted to get an estimate of the scene. Finally, different tokens are assigned with different weightage based on their arrangement, which differentiates between various objects.

### b. Classification:

Extracted features are classified in order to distinguish the objects of interest (say, human) from any other objects present in the videoframe. We have used Nonlinear (cubic) Support Vector Machine (SVM) as the feature classifier. Cubic SVM bundles consider consecutive triples of words for classification. In our case, cubic SVM performed better than all other types of SVM as well as other classifiers. Polynomial kernel for cubic SVM is:

$$K(x, y) = (x^T y + c)^3. \qquad (26)$$

Here $x$ and $y$ are input vector features, calculated from the training samples. A free parameter, $c \geq 0$, is indicating how far the equation is from homogeneity.

### c. Detection of feature of test objects:

In this section, we have followed the similar approach of bag words to extract the textual distribution as feature of the test objects in the video scenes. During classification process, similar feature in different objects can lead to uncertainty in feature assignment. Dealing with uncertainty, implicit shape model has been proposed (ISM). In ISM algorithm, extraction of local features and matching it to the Visual Vocabulary is done using soft matching. At the time of validation of our classification process we used the notion of soft computing which is basically a heuristic interpretation of our matching threshold.

The following equation expresses the contribution of a feature $f$, at location $l$, at position $x$ in the object class $o_n$ with matching visual keywords ($C_i$) indicating its potentiality of belonging to the class $o_n$. Thus, we get:

$$p(o_n, x | f, l) = \sum_i \ p(o_n, x | C_i, l) \ p(C_i | f), \qquad (27)$$

where $p(o_n, x | f, l)$ indicates the probability of feature $f$ at frame location $l$ to belong to the class of $o_n$ at image position $x$.

The weights are populated in continuous 3D weighing region for object position x= (x, y, s). For visual keyword $C_i$, the first term of the right-hand side of (27) indicates the stored occurrence distribution, which is weighted by the second term, the probability of feature $f$, *belonging* to the exact class of $C_i$.

Mean-shift mode estimation with a kernel $K$, along with scale-adaptive kernel, is used to obtain the maxima in this space:

$$\hat{p}(o_n, x) = \frac{1}{V_b(x_s)} \sum_k \sum_j \ p\left(o_n, x_j | f_k, l_k\right) K\left(\frac{x - x_j}{b(x_s)}\right). \qquad (28)$$

Kernel bandwidth is denoted by $b$, and volume is denoted by $V_b$, which are varied over the radius of the kernel. In order to fix the hypothesized interest object, size and scale coordinate $x_s$ is parallelly updated. This strategy makes it easier to deal with partial occlusions and also typically requires fewer training examples.

The pictorial structure model represents any object of interest as collection of parts, connected in parts, and defined by a graph $G = (V, E)$, where the nodes $V = \{v_1, \ldots, v_n\}$ defines the parts and the edges $(v_i, v_j) \in E$ describes the corresponding connections.

$L = \{l_1, \ldots, l_n\}$ be a certain arrangement of part frame locations. Then the matching of the model to a video frame is formulated using an energy minimization function:

$$L^* = \arg\min_L \left( \sum_{i=1}^n m_i(l_i) + \sum_{(v_i, v_j) \in E} d_{i,j}(l_i, l_j) \right). \qquad (29)$$

The matching cost is $m_i(l_i)$ at location $l_i$, with the placing part $v_i$ and deformation cost between two corresponding part locations represented by $d_{i,j}(l_i, l_j)$:

$$d_{i,j}(l_i, l_j) = \left( T_{ij}(l_i) - T_{ji}(l_j) \right)^T M_{ij}^{-1} \left( T_{ij}(l_i) - T_{ji}(l_j) \right), \quad (30)$$

where $d_{i,j}$ is Mahalanobis Distance between transformed locations $T_{ij}(l_i)$ and $T_{ji}(l_j)$, $M_{ij}$ being the diagonal covariance. The root node optimum location stated as:

$$l_1^* = \arg\min_{l_1}\left(m_1(l_1) + \sum_{i=2}^n D_{m_i}(T_{1i}(l_1))\right), \qquad (31)$$

where $D_{m_i}(T_{1i}(l_1))$ represents a sum of the root node's response as a distance-transformed version of each child node.

Thus, we get:

$$l_1^* = \arg\min_{l_1}\Big(m_1(l_1) + \sum_{i=2}^n \min_{l_i} \ m_i(l_i) + ||l_i - T_{1i}(l_1)||^2_{M_{ij}}\Big). \qquad (32)$$

In (Fig. 4), visual word tokenization is explained and furthermore, the visual word occurrence frequency is explained in the following section. It signifies the number of time a particular word vector represents its class of object in varied frames. This is evidently the parameter that decides the characteristics of a particular object.

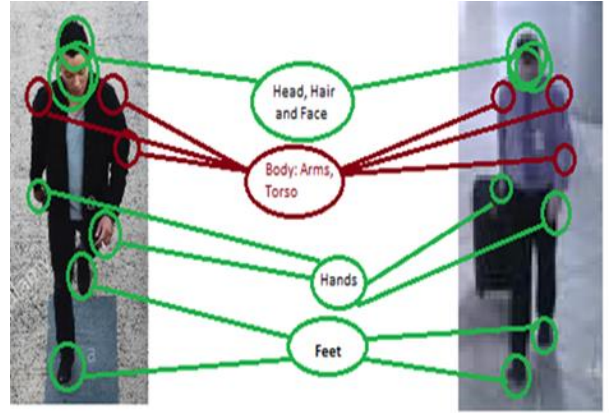

Fig. 4. Tokenization in Bag of Words: Similar feature vectors are grouped as one word, which is later used for matching.

To validate the classification, cross validation approach is applied by randomly splitting the array of objects into train data subsets and test data subsets. Then it creates a confusion matrix for comparing the correct and incorrect classification of features for training of Visual Vocabulary Model. The sum along the principle diagonal of the confusion matrix represents the number of correctly classified objects. The ratio between correctly classified object to the total number of objects to be identified gives the validation accuracy.

### d. Adaption of Discriminative model based on Pyramid Matching Kernel (PMK) approach:

In this paper, we have essentially considered discriminative model for tracking the objects with different appearance (Fig. 9). For further improvement of our validation score by approximating the similarity measures, we are modelling a linear time matching function, represented by the Pyramid Match Kernel (PMK) model to bridge the feature sets to the variable cardinalities. Let the input of a histogram pyramid is $X \in S$ where $\Psi(X) = [H_0(X), \dots, H_{L-1}(X)]$, number of pyramid levels expressed as $L$. The histogram vector of point $X$ is defined by $H_i(X)$.

Similarity between two input set of features $Y$ and $Z$ is expressed as:

$$\kappa_{PMK}(\Psi(Y), \Psi(Z)) = \sum_{i=0}^{L-1} \omega_i \left( I\big(H_i(Y), H_i(Z)\big) - I\big(H_{i-1}(Y), H_{i-1}(Z)\big) \right), \quad (33)$$

where $I\big(H_i(Y), H_i(Z)\big)$ signifies the histogram intersection of two input set of features $Y$ and $Z$ at $i^{\text{th}}$ level of the pyramid.

### E. Region of Interest Tracking

The species inspired PSO framework provides an effective way to track multiple object that are detected and recognized from aforementioned method (visual word features). First, for singular object tracking, following analogies need to be assumed:

- The groundtruth of an object and surrounding region can be considered as ecological properties.
- State space particles correspond to a particular species.
- Each particle's observation likelihood and fitness capability of a particular species is analogous.

For multiple object tracking, these postulates can be easily extended by creating a tracker for each object. These trackers are managed independently. In case of occlusion, support regions of concerning objects may overlap, which implies, the intersectional area between two species are elementary to both. Subsequently, the repulsion and competition among the species arise as both of them aspire to the same resource, the stronger one has higher probability of winning the competition.

During the course of video scene, there may be overlap between two object areas due to occlusion and the related features between them may become ambiguous. To handle this hindrance, we design a multiple-species-based PSO algorithm. The principle idea behind this approach [19], is to divide the groundtruth particles of the object into various species according to the species object numbers and successfully model the relations and the partial visibility among varied species. Detailed description of the species inspired PSO algorithm is briefly described in the following sections.

#### a. Problem Construction:

Let us consider, M number of objects, surrounded with N number of particles, constitute a set $\chi = \{x_{t,k}^{i,n}, i = 1, \dots, N, k = 1, \dots, M\}$, $\boldsymbol{O} = \{o_{t,k}^{i,n}, i = 1, \dots, N, k = 1, \dots, M\}$. Here $t$ is the 2-D translation parameter. Formula of multiple object tracking is as follows:

$$\chi^* = \arg\max_i p(\boldsymbol{O}|\chi) \quad (34)$$

By independently maximizing of the individual observation likelihood, the above optimization may be simplified, in case of no occlusion.

In case of no occlusion, the above optimization may be simplified by maximizing the individual observation likelihood independently (here, we drop the superscript i, n for simplicity):

$$x_{t,k}^* = \arg\max_{x_{t,k}} p(o_{t,k}|x_{t,k}), k = 1, \dots, M \quad (35)$$

#### b. Competition Model:

When different object obscure one another, there is an overlap between corresponding support regions. In these circumstances, the competition between two objects elevates to subjugate the overlapping part (Fig. 5). In order to effectively design the competition phenomenon, the visual problem needs to be merged with the competition process.

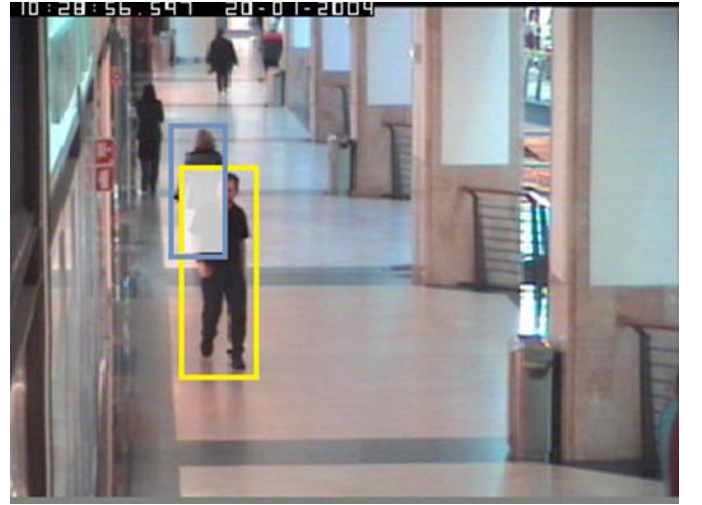

Figure 5. Overlap between two object areas.

To evaluate the fitness value on the overlapping part as the competition ability, the overlapping part is viewed at a whole and projected onto the learned subspace corresponding to each object. We define the power of each object or species in following manner:

$$power^k = p(\hat{o}_{t,k}|x_{t,k}) = exp\left(-\left\|\hat{o}_{t,k} - \widehat{U}_k \widehat{U}_k^T \hat{o}_{t,k}\right\|^2\right), \quad (36)$$

where $k$ and $\widehat{U}_k$ are the overlapping part of the object and its corresponding subspace respectively. In a similar way, the interactive likelihood of object $k_1$ over the overlapping regions can be calculated:

$$\underbrace{p(\hat{o}_{t,k_1}|x_{t,k_1}, x_{t,k_2})}_{interactive\ likelihood} = \frac{power^{k_1}}{\sum_{i=1,2} power^{k_i}}. \quad (37)$$

The mutual likelihood of each species describes the competition ability. Higher the competition ability of a species more like it is to win the competition. It means that the species which won the competition is more likely to be of the object that was occluding the other object species involved.

#### c. Annealed Gaussian Based PSO (AGPSO)

An annealed Gaussian based PSO algorithm [21] is considered in this paper, as in conventional PSO requires careful and fine tuning of various parameters. In this algorithm, the particles and corresponding velocities are updated in below mentioned manner:

$$v^{i,n+1} = |r_1|(p^i - x^{i,n}) + |r_2|(g - x^{i,n}) + \epsilon \quad (38)$$

$$x^{i,n+i} = x^{i,n} + v^{i,n+1}, \qquad (39)$$

where $|r_1|$ and $|r_2|$ being the absolute values of the samples from Gaussian probability distribution N (0, 1). This is zero-mean Gaussian disturbance that stops the algorithm from getting trapped in local optima. With the help of adaptive simulated annealing, the covariance matrix of $\epsilon$ is changed [34]:

$$\Sigma_\epsilon = \Sigma \, e^{-cn}. \qquad (40)$$

Here, a transition distribution is predefined and $\Sigma$ is its covariance matrix, annealing constant $c$, and iteration constant $n$. The components in $\Sigma$ decrease in proportion to the iteration number which results in a fast rate of convergence. When $k_1$ and $k_2$ occlude each other at time t, a repulsion force is added to the evolution process of particles, and subsequently the iteration step for $k_1$ becomes as follows:

$$v_{t,k_1}^{i,n+1} = |r_1| \big( p_{t,k_1}^i - x_{t,k_1}^{i,n} \big) + |r_2| \big( g_{t,k_1} - x_{t,k_1}^{i,n} \big) +$$
$$|r_3| F_{k_2,k_1} + \epsilon \qquad (41)$$
$$x_{t,k_1}^{i,n+1} = x_{t,k_1}^{i,n} + v_{t,k_1}^{i,n+1}, \qquad (42)$$

where the parameter $r_3$ is Gaussian random number sampling from N (0, 1). The third term on the right-hand side of the above equation depicts the shared effect between object $k_2$ and $k_1$. In other words, the competition phenomenon on the observation level has been modelled in this paper. Also, the competition model of state space has been modelled to drive the evolution process of the species in the right direction.

#### d. Updating of the Appearance Model Selectively

In most of the tracking algorithms [24], [29], appearance models are not updated during occlusion. However, the appearance of the object under occlusion may change, and that can cause the tracker to fail to recapture the object appearance after it is not occluded anymore. A selective updating algorithm is implemented to cope with the appearance changes during occlusion: pixels belonging to the visual part of the objects are cumulatively updated in the normal way and pixels that are part of the overlapping region (Fig. 5) are projected onto the subsequent subspace of each object. Then the errors due to the reconstruction are calculated. If this error is smaller than a predefined threshold for pixels inside the overlapping area, then it is again updated in the subsequent subspace.

Due to this careful modelling of the updating strategy, the appearance changes can be easily accommodated, allowing more persistent tracking throughout the video stream.

### IV. EXPERIMENTAL RESULTS AND ANALYSIS

Proposed method is evaluated on benchmark datasets and compared with the existing state-of-the art tracking algorithms. Brief description of the experimental datasets is shown followed by the experimental parameter settings and analysis.

All the tests were done on an Intel 5th Gen core i7, 2.10 GHz processor with 6 Gigabytes of RAM and 2 Gigabytes NVIDIA GeForce GPU and the algorithm was implemented using MATLAB'16 development tool.

#### A. Experimental Datasets

We evaluate the proposed algorithm on benchmark TB100 sequences, namely, iLIDS [4], VIVID [46], Walking2 [50] and Woman [40]. In the following section, we discuss the various attributes of the aforementioned datasets.

The primary challenges in the iLIDS dataset [4] (imagery Library for Intelligent Detection Systems) is the Scale Variation (SV), In Plane Rotation (IPR), Occlusion (OCC), Low Resolution (LR) and Illumination Variation (IV). This video consists of a total number of four people walking to, from and across the camera view, with the camera mounted at an isometric angle, where only one man carries a trolley. The video is 10 minutes long, at 25 frames per second, with each frame having a dimension of 720x570 pixels, which we have scaled down to 180x144 for low resolution analysis.

The woman dataset [40] is much more challenging, as one has to handle Illumination variation, Scale variation, Occlusion, Deformation (DEF), Motion Blur (MB), Fast Motion (FM) and Out Plane Rotation (OPR). It has 597 video frames of 352x288 resolution. Here the camera view follows a woman walking past several cars.

The VIVID dataset [46] is part of DARPA VIVID program consisting 9 video sequences. Here we deal with one of those nine videos, namely, RedTeam. The video contains challenges such as; Scale Variation, Occlusion, In Plane Rotation, Out Plane Rotation and Low Resolution. This is an aerial footage of a car driving on a straight road then turning a corner at the end forming long shadow cast of the object. Therefore, shadow removal gives a better result in tracking.

Walking2 dataset [50] contains difficulties like Scale Variation, Occlusion and Low Resolution. The video contains 500 frames, each of dimension 384x288 pixels. The video is of some people walking down the corridor of an office interior. This video is in many ways similar to the iLIDS dataset. In the following section, we demonstrate the accuracy of our algorithm and susceptibility in handling all the challenges offered by these benchmark datasets.

#### B. Experimental Setting

All the aforementioned datasets consist of RGB color channels, which are more memory extensive for image processing. So, we temporarily convert them to grayscale using the equation: $gray = 0.299R + 0.587G + 0.114B$.

Furthermore, we update the background model with the process mentioned in (14) (15) and (16) of our proposed method.

Then we process each frame with their original RGB color channels and reconstruct a new image by feeding only the normalized $R$ and $G$ color channel values. This normalized image removes the shadow component from the image, which essentially, helps in accurate tracking. After this step, we convert this image into a binary image through background subtraction [Figure 6.(b)]. We dilate the image by employing some morphological operations. However, this image does not preserve the edges satisfactorily enough. Thus, we extract the binary image with edges intact, by simple background subtraction [Figure 6. (c)]. Then we pointwise multiply these images to obtain the complete shadow free image [Figure6. (d)] with edges preserved.

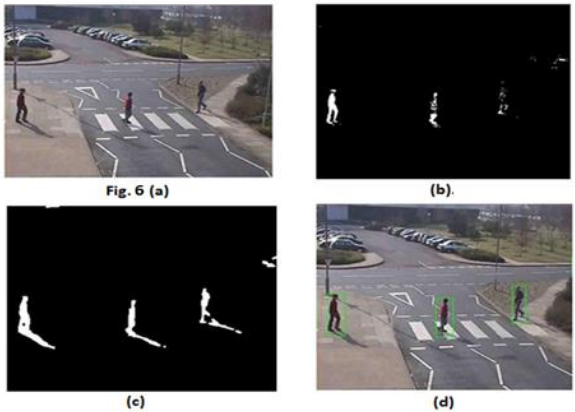

Fig. 6. (a) Original Image, (b) Color Invariant Image, (c) Simple Background Subtracted Image, (d) Reconstructed Image

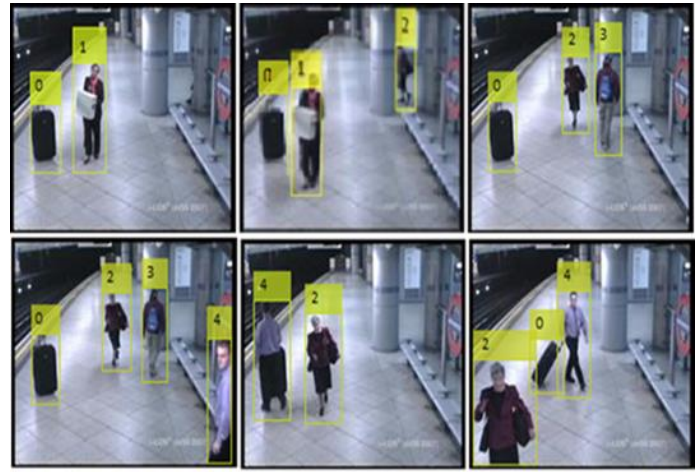

Fig. 8. Tracking Results

For training, we consider the domain with train data sets and we use Visual Vocabulary features. We're extracting key points and calculating the probable occurrences in the subsequent frames (Fig. 7). Continuing with this process, we extract all the key points of certain domain oriented objects from different angle-posture-view high definition picture datasets for better accuracy.

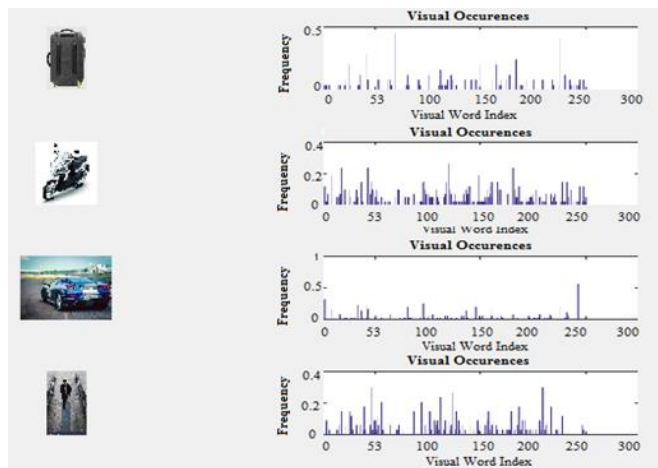

Fig. 7. Training Results

Visual Vocabulary using Cubic SVM based classification covers two other parameters, namely domain detection of the scene and object identification for further video frames, which are independent to camera axis orientation, camera background relationship and surroundings. It provides recognition accuracy of roughly 93.3%.

Then we use the extracted and trained features to the Species inspired PSO for accurate and content aware tracking. Following this process, as we can see in (Fig. 8), the unattended luggage is also recognized. Here the luggage is a rigid object which remains stationary for most of the video and when it is moved by any human, it forms a connected blob with that human. Normally, in such cases, PSO cannot track successfully, but because of the feature driven input to the Species inspired PSO, we get the results shown below.

## C. Analysis and Evaluation

We test our algorithm on the dataset mentioned above with the aforementioned setting. The confusion matrix (Fig. 10) shows that we have reached up to 85.33% accuracy. Here, we consider INRIA Person Dataset for training of our Visual Vocabulary Model from various angle and postures for accurate detection and recognition, as portrayed in Fig. 9.

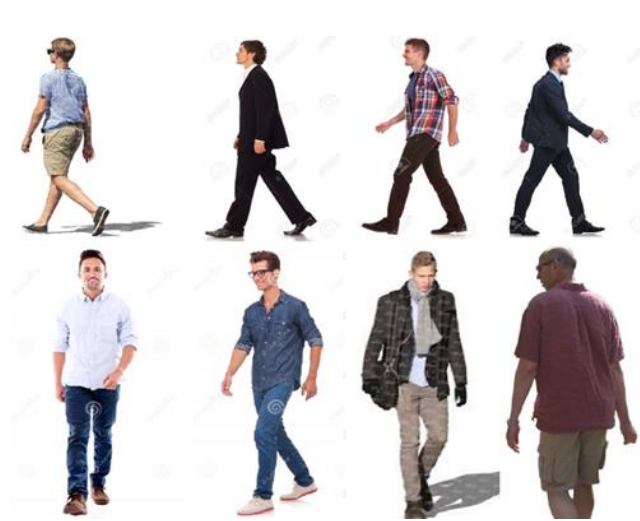

Fig. 9. Training image montage: Moderately high-resolution pictures of the object of interest, preferably from different angles

More training always leads to higher accuracy. Our training data here consisted of partial photos of train that had almost similar dimension like cars. That is why our classifier has maximum confusion in this situation, as reflected by the confusion matrix.

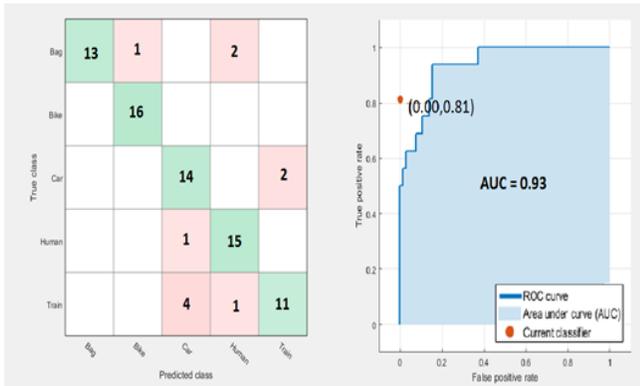

Fig. 10. Confusion Matrix and ROC Curve

Other plots like, ROC curve and scatter point data show the classifier performance and the plot between True Positive Rate and False Positive Rate. Prediction model curve shows the structure containing a classification object and a function for prediction. This structure allows to make predictions for data models that include principal component analysis (PCA).

Applying the extracted features to the test datasets, using Visual Key points, prediction of new objects of interest in the video sequence is done. Subsequently, using the trained model as a reference to recognize newly arrived objects, as shown in the (Fig. 11), with a validation accuracy of roughly 93.13%.

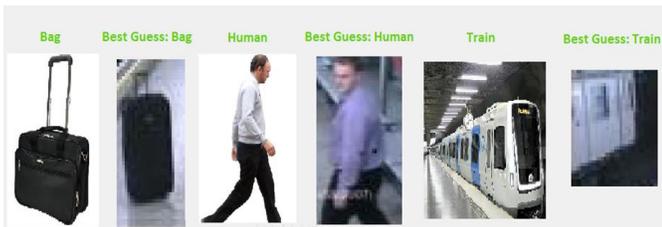

Fig. 11. Prediction on Test Data

Train models using machine learning learners are applied in the video sequences and the algorithm predicts with recognition, the objects of interests, present in the consequent video frames, as stated above.

Domain recognition of the test sequence also being predicted by the probability distribution of presence of objects [i.e. Visual Key Points] in the scene, as shown in the [Fig. 12]

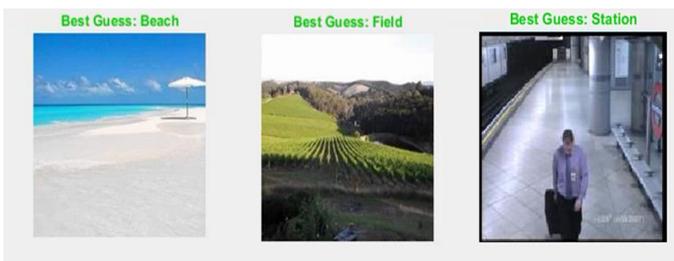

Fig. 12. Domain Recognition

Now, we compare the detection accuracy of our algorithm with different benchmark methods as shown in Table 1, 2, 3, 4.

Next, a comparative analysis of the Frame Per Second (FPS), provided in [5], is demonstrated in Table 5. We implement our algorithm on the VIVID dataset [46], and compare it with various state-of-the-art methods to obtain the necessary data for Table 5.

Table 1. Detection Accuracy on iLIDS Dataset [4]

| Approach | Year | Accuracy |
|---|---|---|
| ASLA [11] | 2012 | 79.24% |
| DFT [12] | 2012 | 81.21% |
| IVT [32] | 2008 | 82.36% |
| MIL [17] | 2009 | 84.29% |
| PCOM [6] | 2014 | 79.87% |
| LSS [8] | 2013 | 78.39% |
| **Proposed method** | | **91.3%** |

Table 2. Detection Accuracy on VIVID Dataset [46]

| Approach | Year | Accuracy |
|---|---|---|
| ASLA [11] | 2012 | 86.3% |
| DFT [12] | 2012 | 85.3% |
| IVT [32] | 2008 | 88.9% |
| MIL [17] | 2009 | 90.2% |
| PCOM [6] | 2014 | 86.7% |
| LSS [8] | 2013 | 87.2% |
| **Proposed method** | | **92.7%** |

Table 3. Detection Accuracy on Walking2 Dataset [50]

| Approach | Year | Accuracy |
|---|---|---|
| ASLA [11] | 2012 | 88.2% |
| DFT [12] | 2012 | 87.9% |
| IVT [32] | 2008 | 89.8% |
| MIL [17] | 2009 | 90.2% |
| PCOM [6] | 2014 | 88.3% |
| LSS [8] | 2013 | 90.7% |
| **Proposed method** | | **93.5%** |

Table 4. Detection Accuracy on Woman Dataset [40]

| Approach | Year | Accuracy |
|---|---|---|
| ASLA [11] | 2012 | 87.8% |
| DFT [12] | 2012 | 88.2% |
| IVT [32] | 2008 | 90.9% |
| MIL [17] | 2009 | 91.2% |
| PCOM [6] | 2014 | 88.4% |
| LSS [8] | 2013 | 89.1% |
| **Proposed method** | | **93.7%** |

Table 5. Evaluated Tracking Algorithms

| Trackers | Representation | | | | | | | | | | Code | | | |
|---|---|---|---|---|---|---|---|---|---|---|---|---|---|---|
| | Local | Template | Color | Histogram | Subspace | Sparse | Binary or Haar | Discriminative | Generative | Model Update | C | Matlab | FPS | Published |
| ASLA | ✓ | ✓ | | | ✓ | ✓ | | | ✓ | ✓ | ✓ | ✓ | 8.5 | '12 |
| BSBT | | | | | | | H | ✓ | | ✓ | ✓ | | 7.0 | '09 |
| CXT | | | | | | | B | ✓ | | ✓ | ✓ | | 15.3 | '11 |
| DFT | ✓ | ✓ | | | | | | | ✓ | ✓ | | ✓ | 13.2 | '12 |
| IVT | | ✓ | | | ✓ | | | | ✓ | ✓ | ✓ | ✓ | 33.4 | '08 |
| LIAPG | | ✓ | | | ✓ | ✓ | | | ✓ | ✓ | | ✓ | 2.0 | '12 |
| LOT | ✓ | | ✓ | | | | | | ✓ | ✓ | | ✓ | 0.7 | '12 |
| LSS | ✓ | ✓ | | | ✓ | ✓ | | | ✓ | ✓ | | ✓ | 15 | '13 |
| MIL | | | | | | | H | ✓ | | ✓ | ✓ | | 38.1 | '09 |
| MTT | | ✓ | | | | ✓ | | | ✓ | ✓ | | ✓ | 1.0 | '12 |
| ORIA | | ✓ | | | ✓ | | H | ✓ | | ✓ | ✓ | | 20.2 | '11 |
| PCOM | | ✓ | | | ✓ | ✓ | | | ✓ | ✓ | ✓ | | 20 | '14 |
| SMS | | | ✓ | ✓ | | | | | ✓ | | ✓ | | 19.2 | '03 |
| **Proposed method** | ✓ | ✓ | ✓ | ✓ | | ✓ | ✓ | | | ✓ | | ✓ | **29** | — |

In all these cases, our algorithm performs competitively better than all the popular existing approaches. This is due to the fact that other algorithms are effective to deal with certain challenges offered by the datasets, but they are not susceptible enough to cope up with all the challenges of datasets, as stated earlier in section IV (A). Table 6 represent the comparative performance of different trackers with our proposed algorithm. In Table 6, each entry has a numerator term which represents tracking score and the denominator term represents the false positive in tracking.

A careful inspection of Table. 6 reveals the fact that in all the cases our algorithm performs better than the existing ones. Fig. 14 graphically depicts the above stated fact about the performance measure which indicates the proposed algorithm's flexibility in adapting with real-life challenges.

Fig. 13 further shows a comparative study on the performance of our algorithm with respect to tracking accuracy and the size of Visual Vocabulary Model. Thus, the proposed algorithm performs very satisfactorily over different challenging attributes of different images and illumination variations of the video frames.

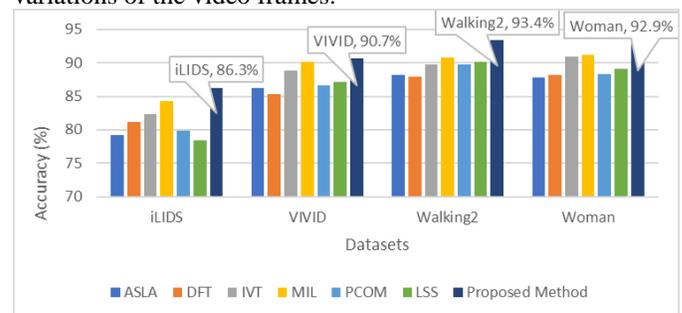

Fig. 13. Graphical Representation of Accuracy Analysis

As it can be seen in (Fig. 14), our algorithm performs best on Walking2 datasets, which is expected because of fewer number of challenging attributes. It performs well on iLIDS and VIVID datasets despite the low-resolution video frames.

Table 6. Performance of different algorithms on different attributes

| Trackers / Features | ASLA [11] | DFT [12] | IVT [32] | MIL [17] | PCOM [6] | LSS [8] | Proposed method |
|---|---|---|---|---|---|---|---|
| Scale Variation (SV) | 54.0 / 3.9 | 47.9 / 5.9 | 47.1 / 5.3 | 44.5 / 6.5 | 44.8/5.7 | 48.5 / 5.3 | 58.0/4.1 |
| In Plane Rotation (IPR) | 52.1 / 4.1 | 50.7 / 5.1 | 46.4 / 5.3 | 45.7 / 5.9 | 43.7/5.9 | 47.1 / 5.5 | 50.7/3.9 |
| Occlusion (OCC) | 56.0 / 3.8 | 52.7 / 5.1 | 49.3 / 5.1 | 47.6 / 5.8 | 47.4/5.5 | 51.2 / 5.1 | 60.9/1.6 |
| Illumination Variation (IV) | 59.6 / 3.0 | 53.0 / 4.7 | 51.2 / 4.8 | 47.1 / 5.6 | 47.8/5.8 | 51.4 / 5.2 | 59.9/2.7 |

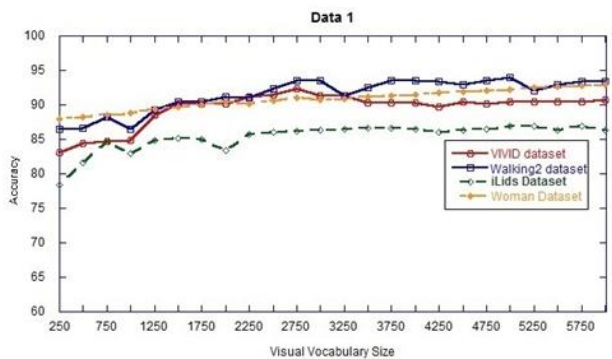

Fig. 14. Accuracy vs Visual Vocabulary of the proposed algorithm, plotted on 4 datasets.

## V. CONCLUSION

This paper presents object detection, recognition of the detected objects based on Visual Vocabulary Model and tracking of the recognized objects using Species inspired PSO. We train different objects separately in several images with multiple aspects and camera viewpoints to find the best key word points for recognition. Subsequently, we verify the extracted features of the train images. These key word points are applied to the regions based on visual feature point analysis. The comparative analysis is done using visual key word points. We present similarity measures using PMK approach [Section IV, D(d)] for feature matching. The object is satisfactorily detected. After detection of the object, the recognition of the specific object of our interest is done in section IV (D). Finally, the features of the recognized objects are tracked by the Species inspired PSO, which can also efficiently handle the tracking under partial occlusions as shown in Fig 8. The performance measure of the proposed algorithm is done with respect to available benchmark data [4, 46, 50, 40] and we obtain very satisfactory and competitive results. In future, we have a plan to modify the detection and recognition scheme based on the theory of SP (Simplicity & Power) Intelligence.